\documentclass[runningheads]{llncs}
\usepackage[T1]{fontenc}

\usepackage{graphicx}

\makeatletter
\g@addto@macro\normalsize{%
  \setlength\abovedisplayskip{4pt plus 2pt minus 2pt}%
  \setlength\belowdisplayskip{4pt plus 2pt minus 2pt}%
  \setlength\abovedisplayshortskip{2pt plus 2pt minus 1pt}%
  \setlength\belowdisplayshortskip{2pt plus 2pt minus 1pt}%
}
\makeatother

\usepackage{algorithm}
\usepackage{algpseudocode}

\usepackage{amsmath,amssymb}

\usepackage{cite}
\usepackage{url}
\usepackage[hidelinks]{hyperref}

\usepackage{xcolor}
\usepackage{microtype}

\usepackage{booktabs}
\usepackage{multirow}
\usepackage{array}
\usepackage{makecell}
\usepackage{tabularx}
\begin{document}
\title{MTF-Net: Multi-Modal Temporal Feature Fusion Network for Pedestrian Intention Prediction}
%
%
\author{Md Mahfuzur Rahman\inst{1}\orcidID{0009-0003-9007-4395} \and
Pengzhan Zhou\inst{1}\orcidID{0000-0002-8796-5969} \and
A. F. M. Abdun Noor\inst{2}\orcidID{0009-0008-6954-5523} \and
Md Imam Ahasan\inst{1}\orcidID{0009-0009-8407-2071} \and
Md Mustafizur Rahman\inst{1}\orcidID{0009-0008-6218-0605} \and
Fang Qu\inst{1}\orcidID{0009-0003-5669-2923}}
\authorrunning{Mahfuzur et al.}
%
\institute{College of Computer Science, Chongqing University, Chongqing, China \\
\email{pzzhou@cqu.edu.cn}
\and
Department of Software Engineering, Daffodil International University, Bangladesh
}
\maketitle              
\begin{abstract}
Accurately predicting pedestrian intentions is crucial for ensuring safe and proactive interaction between autonomous vehicles and pedestrians. However, existing approaches often depend on architectures that either model temporal dependencies within individual modalities or fuse modalities only at coarse semantic levels. To address these limitations, we propose MTF-Net, a novel Multi-Modal Temporal Feature Fusion Network that jointly models kinematic, appearance, and contextual cues for pedestrian intention prediction. MTF-Net integrates four complementary modalities-bounding-box dynamics, human pose keypoints, local context, and scene-level semantics within a recurrent fusion framework enhanced by gated linear units (GLUs). These GLU-based modules adaptively regulate cross-modal information flow, enabling interpretable and efficient feature interaction across temporal scales. Through three dedicated temporal encoding branches and an attention-guided fusion head, the proposed model robustly anticipates pedestrian crossing intentions several frames before they occur. Extensive evaluations on the PIE and JAAD benchmarks demonstrate that MTF-Net surpasses recent transformer- and graph-based models, achieving up to $0.95$ AUC on PIE and $0.94$ AUC on JAAD, while maintaining real-time performance. The results highlight that reliable pedestrian intention prediction arises from principled multi-modal fusion rather than excessive architectural complexity.

\end{abstract}
\section{Introduction}
\label{sec:intro}
Pedestrians account for nearly 23\% of global road traffic fatalities~\cite{world2019global}, underscoring the critical importance of accurately predicting pedestrian intentions for improving the safety of advanced driver assistance systems (ADAS) and autonomous vehicles (AVs). For autonomous systems to interact effectively with pedestrians, they must not only prevent collisions but also facilitate safe crossings and maintain smooth traffic flow, particularly in dense and dynamic urban environments~\cite{katyal2020intent,10160660}. In unstructured or unsignalized environments, pedestrian motion becomes highly uncertain~\cite{parikh2024idd}. Ego-vehicles in such environments must anticipate changes in pedestrian motion and intent in real time, making reliable and interpretable intention prediction crucial for safe and socially aware navigation. Predicting such intentions requires a fine-grained understanding of motion dynamics, social interactions, and contextual semantics, which extend beyond conventional trajectory prediction or detection tasks. Despite recent advances in trajectory forecasting~\cite{shi2021sgcn, li2022graphbased}, pedestrian detection~\cite{khan2023localized, cheng2025lightweight,song2023optimal, liu2023vlpd}, and animation modeling~\cite{rempe2023trace, wang2024pacer}, the explicit prediction of pedestrian crossing intention remains largely underexplored. 
\begin{figure}[!ht]
    \centering
    \includegraphics[width=1\linewidth]{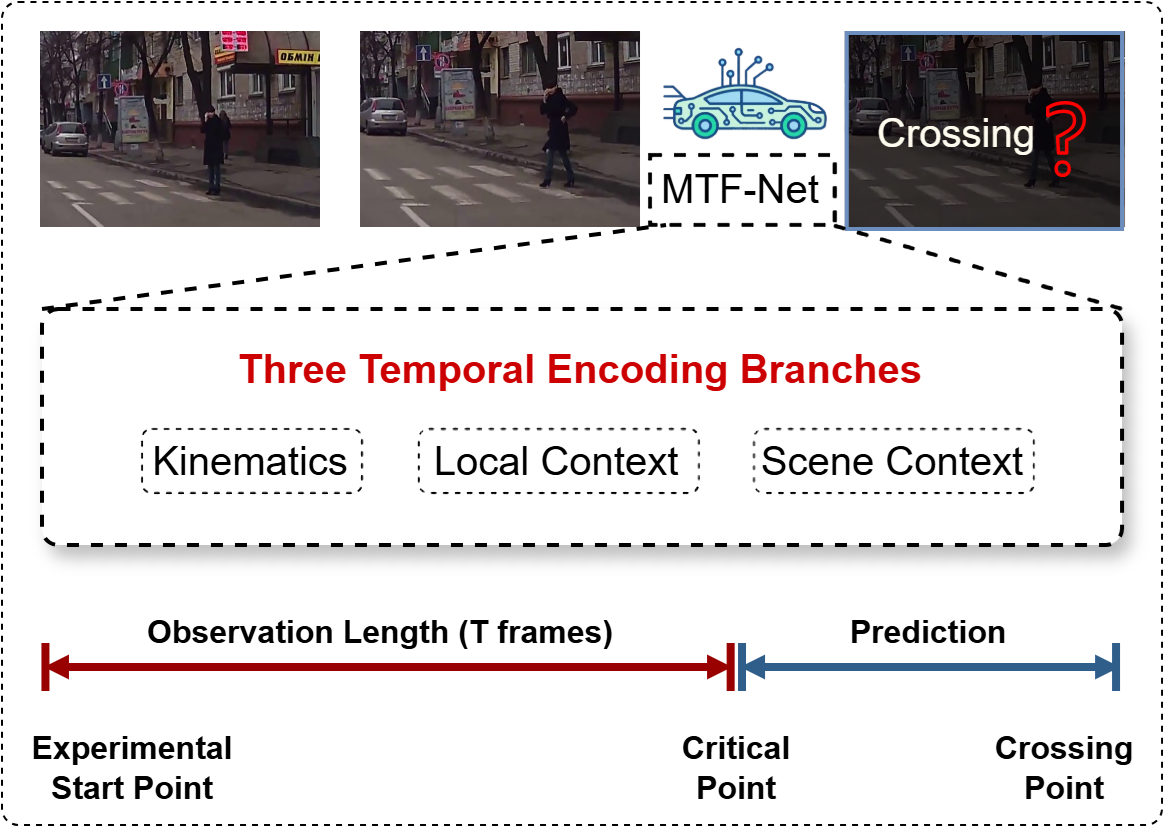}
    \caption{Conceptual overview of MTF-Net for pedestrian intention prediction.The model uses multimodal features kinematics, local context, and scene context through three temporal encoding branches. These features are fused via GLUs to anticipate pedestrian crossing intention.}
    \label{fig:concept}
\end{figure}
A wide range of techniques has been investigated to infer and predict pedestrian crossing intentions, spanning convolutional neural networks (CNNs)~\cite{razali2021pedestrian,zhang2022st,GeGLUNet,azarmi2025pip}, recurrent neural networks (RNNs)~\cite{rasouli2020pedestrian,kotseruba2020they,rasouli2022multi}, graph convolutional networks (GCNs)~\cite{cadena2022pedestrian,chen2021visual}, and Transformers~\cite{s21175694,zhou2023pit,zhang2025prior,sharma2025predicting}. These models typically leverage visual cues such as body pose~\cite{yang2021pedestrian,ahmed2023multi}, spatial temporal relationships~\cite{liu2020spatiotemporal}, and pedestrian vehicle interactions~\cite{xu2024pedestrian}. Despite strong performance, many struggle with generalization, contextual understanding, and causal reasoning in highly dynamic scenes~\cite{zhang2024causal}. 

To address the challenges of accurate and efficient pedestrian intention prediction in complex urban scenarios, we propose \textbf{MTF-Net}, a \textit{Multi-modal Temporal Feature Fusion Network} that jointly reasons over motion, appearance, and contextual cues. As illustrated in Fig.~\ref{fig:concept}, MTF-Net encodes kinematic, local, and scene-level representations through three dedicated temporal branches and fuses them via gated linear units (GLUs) to regulate cross-modal interactions. This design bridges the gap between fine-grained temporal reasoning and robust multi-modal feature alignment, two complementary aspects that are often addressed independently in prior research. Collectively, this study contributes to bridging the gap between trajectory prediction and behavioral understanding, offering a proactive approach to modeling pedestrian intention for autonomous perception systems. The main contributions of our work are summarized as follows:
\begin{itemize}
    \item We introduce a novel temporal fusion framework that integrates four complementary modalities bounding-box dynamics, human pose keypoints, pedestrian-centered local context, and scene-level semantic context within a unified recurrent structure.
    \item Dedicated GRU-based encoders are developed for kinematic, local, and scene features, each enhanced with attention mechanisms for adaptive temporal reasoning and context refinement.
    \item We employ a robust training strategy, achieving an excellent balance between accuracy and efficiency suitable for real-time deployment.
\end{itemize}

\section{Related Work}
\label{sec:related}
\subsection{Pedestrian Intention Prediction (PIP)} 
Recent studies~\cite{munir2025pedestrian,yang2023dpcian,yang2024faster,ham2023cipf,ling2024pedast,chen2024pedestrian,cadena2022pedestrian,xu2024pedestrian,zhou2023pit,zhang2022st,zhang2023trep} have focused primarily on binary crossing-noncrossing prediction using onboard vehicle cameras, achieving remarkable accuracy. Another study ~\cite{masoud2024machine} has highlighted the role of factors such as speed variability and red-light violations in shaping pedestrian vehicle interaction dynamics at intersections. These works illustrate the growing maturity of PIP as a key perception component for autonomous systems. The evolution of PIP frameworks has progressed from single-frame CNN-based estimators~\cite{rasouli2017they} to sophisticated architectures broadly categorized into spatio-temporal modeling and feature fusion. Spatio-temporal models typically employ CNNs for visual encoding followed by RNNs for temporal reasoning~\cite{liu2020spatiotemporal,lorenzo2020rnn,kotseruba2020they}, as exemplified by PIE~\cite{rasouli2019pie}, which integrates ConvLSTMs for sequential analysis. Other studies leverage 3D CNNs such as 3D DenseNet~\cite{cui2025geometry,saleh2020spatio}, C3D~\cite{tran2015learning}, or I3D~\cite{carreira2017quo}, and employ graph-based reasoning over human pose structures~\cite{cadena2019pedestrian}. 
\subsection{Multi-modal Feature Fusion Methods}
Several studies aim to integrate heterogeneous cues across modalities to improve robustness and interpretability. SF-GRU~\cite{rasouli2020pedestrian} performs hierarchical fusion across semantic levels, while PCPA~\cite{kotseruba2021benchmark} employs RNN encoders with temporal and modality attention to dynamically reweight feature contributions. To enhance environmental comprehension, MaskPCPA~\cite{yang2022predicting} and subsequent extensions~\cite{rasouli2022multi} incorporate segmentation maps through attention-based fusion mechanisms. More recently, Transformer-based architectures~\cite{achaji2022attention,rosin2025ecam,zhou2023pit,s21175694,zhang2023trep} have demonstrated strong potential for multi-modal integration, yet remain constrained by high data requirements and computational complexity. In contrast, RNN-based fusion frameworks continue to offer competitive accuracy with improved interpretability, an essential property for safety-critical applications such as autonomous driving.

\begin{figure*}[!ht]
    \centering
    \includegraphics[width=1\linewidth]{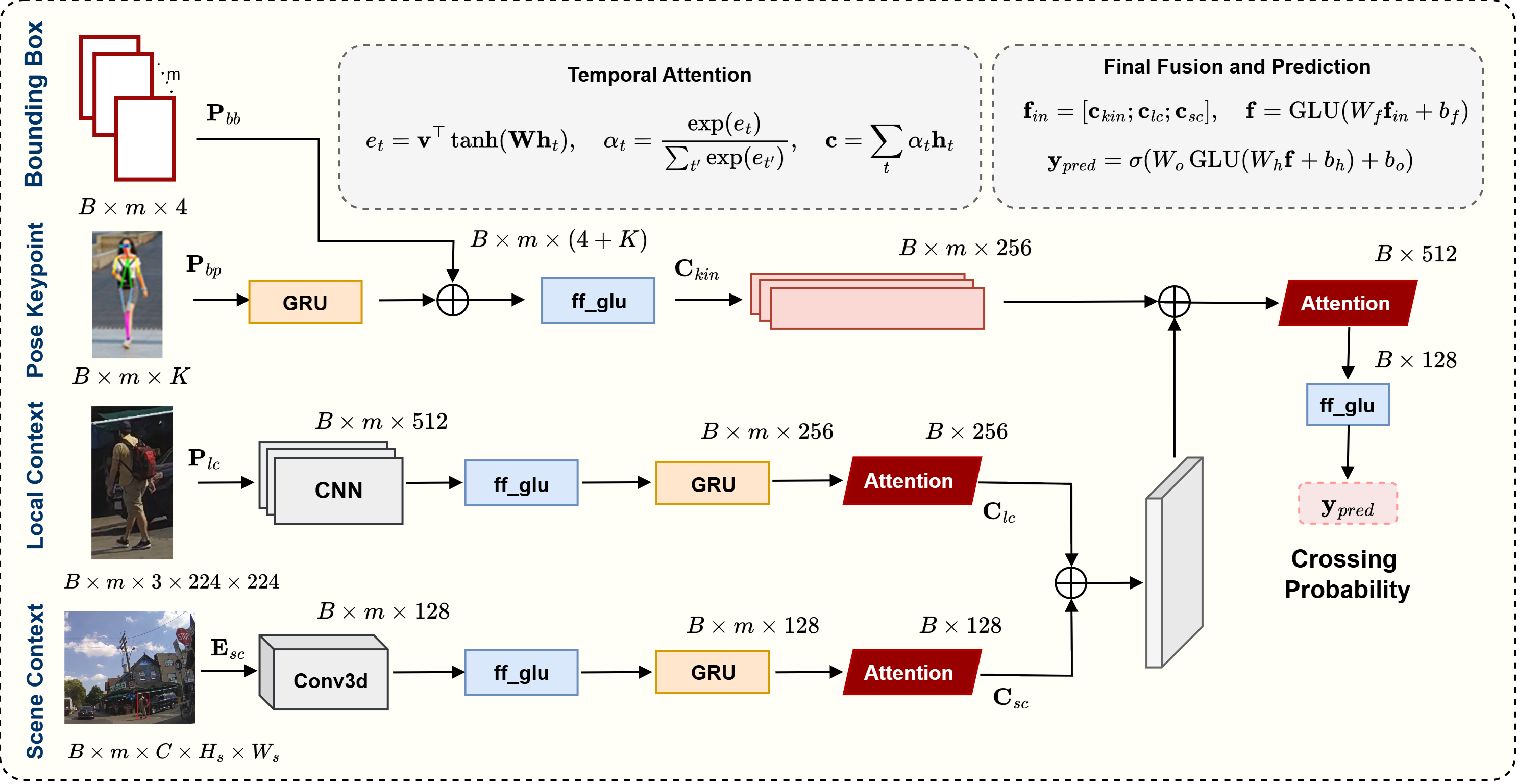}
    \caption{Overview of the proposed MTF-Net architecture. Four input modalities bounding box ($P_{bb}$), pose keypoints ($P_{bp}$), local context ($P_{lc}$), and scene context ($E_{sc}$) are processed through three temporal encoding branches with GRU and attention. GLUs regulate cross-modal fusion, and the attention-guided head predicts pedestrian crossing intention.}
    \label{fig:MTF_Net}
\end{figure*}
\vspace{-12pt}
\section{Methodology}
\subsection{Problem Definition}

Pedestrian intention prediction aims to anticipate whether a pedestrian observed in a video sequence will initiate a crossing action within a predefined temporal horizon. Formally, let an observation window of $T$ frames be represented as
\begin{equation}
X = \{x_t\}_{t=1}^{T}, \quad x_t = [P_{bb,t},\, P_{bp,t},\, P_{lc,t},\, E_{sc,t}].
\end{equation}
where $x_t$ denotes the multimodal observation of a pedestrian at time step $t$. Each $x_t$ is composed of four feature modalities: 
bounding box coordinates $P_{bb,t} \in \mathbb{R}^{4}$, pose keypoints $P_{bp,t} \in \mathbb{R}^{K}$,
local context image patch $P_{lc,t} \in \mathbb{R}^{3\times H \times W}$, and scene context embedding $E_{sc,t} \in \mathbb{R}^{C\times H_s \times W_s}$. Given the observation window $X$, the model $\mathcal{F}_\theta(\cdot)$ parameterized by $\theta$ learns to estimate the probability that the pedestrian will begin crossing within the next $\tau$ seconds:
\begin{equation}
\hat{y} = \mathcal{F}_{\theta}(X) = \mathbb{P}\!\left( y = 1 \mid X \right), \quad \hat{y} \in [0,1].
\end{equation}
where $y \in \{0,1\}$ denotes the ground-truth intention label ($y=1$ for crossing and $y=0$ otherwise). In practice, the observation length $T$ and the prediction horizon $\tau$ are dataset-dependent (typically $T\!\in\!\{10,15\}$ frames and $\tau\!\in\!\{1,2,3,4\}$ seconds). The model thus learns to infer latent spatio-temporal dependencies that correlate multi-modal visual cues with the pedestrian’s forthcoming crossing intention.

\subsection{Multi-Modal Feature Extraction}
Pedestrian behavior is governed by a combination of motion, appearance, and environmental context. 
To capture these complementary cues, MTF-Net processes four synchronized feature modalities: 
bounding box coordinates ($P_{bb}$), body pose keypoints ($P_{bp}$), local context ($P_{lc}$), and global scene semantics ($E_{sc}$). Fig.~\ref{fig:MTF_Net} illustrates the overall MTF-Net architecture, highlighting the multi-modal feature streams and their temporal alignment prior to fusion. All features are extracted using pretrained computer vision models and temporally aligned over the observation window. \textbf{Bounding Box ($P_{bb}$).}
Pedestrian bounding boxes are detected in each frame using YOLOv11~\cite{Khanam2024YOLOv11AO}, 
which provides precise object localization with real-time efficiency. 
For frame $t$, the bounding box is represented as
\begin{equation}
P_{bb,t} = [\,x_{1,t},\, y_{1,t},\, x_{2,t},\, y_{2,t}\,] \in \mathbb{R}^{4}.
\end{equation}
and normalized by the image width ($W$) and height ($H$):
\begin{equation}
P_{bb,t}
= \Big[\, \tfrac{x_{1,t}}{W},\, \tfrac{y_{1,t}}{H},\, \tfrac{x_{2,t}}{W},\, \tfrac{y_{2,t}}{H} \,\Big]
\in [0,1]^4.
\end{equation}
The resulting sequence $P_{bb} \in \mathbb{R}^{T\times 4}$ encodes coarse spatial displacement and scale variation across time, providing low-level kinematic cues. \textbf{Pose Keypoints ($P_{bp}$).}
Fine-grained body dynamics are extracted via YOLO-Pose~\cite{Maji2022YOLOPoseEY}, 
which estimates $K$ two-dimensional keypoints for each detected pedestrian. 
For frame $t$, the normalized pose vector is
\begin{equation}
P_{bp,t}
= \big\{\, (\tfrac{u_{k,t}}{W},\, \tfrac{v_{k,t}}{H}) \,\big\}_{k=1}^{K}
\in \mathbb{R}^{2K}.
\end{equation}
Stacking these across all frames forms
\begin{equation}
P_{bp} = [\, P_{bp,1},\, P_{bp,2},\, \ldots,\, P_{bp,T} \,] \in \mathbb{R}^{T \times 2K}.
\end{equation}
which captures temporal motion of limbs and orientation shifts preceding a crossing event. \textbf{Local Context ($P_{lc}$).} The visual appearance surrounding the pedestrian often reveals intent through gaze, body orientation, or walking direction. Using the YOLO-detected bounding box, a local RGB crop ($I$) is extracted from each frame and enlarged by a factor ($\alpha=1.5$) to include near-field context. 
Each crop is resized to a fixed spatial size $(H_c, W_c)$, yielding
\begin{equation}
P_{lc,t} = \text{Resize}\!\big( I_t[\alpha P_{bb,t}] \big)
\in \mathbb{R}^{3 \times H_c \times W_c}.
\end{equation}
where $I_t$ denotes the raw frame at time $t$. 
This modality focuses on local context for subsequent CNN encoding. \textbf{Scene Context ($E_{sc}$).} Understanding the broader traffic environment is essential for intention inference. 
We employ the MiDaS depth estimator~\cite{Ranftl2019TowardsRM} and Mask2Former segmentation network~\cite{Cheng2021MaskedattentionMT} to obtain per-frame depth and panoptic maps, which are fused into multi-channel semantic tensors:
\begin{equation}
E_{sc,t} = f_{\text{scene}}\!\left( I_t \right) \in \mathbb{R}^{C \times H_s \times W_s}.
\end{equation}
This representation encodes road geometry, vehicle proximity, and crosswalk visibility, allowing the model to reason about affordances and environmental cues. \textbf{Temporal Alignment.} All four modalities are synchronized along the observation window 
$X = \{x_t\}_{t=1}^{T}$ such that
\begin{equation}
x_t = [\, P_{bb,t},\, P_{bp,t},\, P_{lc,t},\, E_{sc,t} \,] \in \mathbb{R}^{d_x}.
\end{equation}
ensuring that multimodal observations correspond to the same timestamp $t$. 
Each modality is cached for efficient loading during training and evaluation.

\begin{algorithm}[!ht]
\caption{MTF-Net Training with Multi-Branch Temporal Encoding and GLU Fusion}
\label{alg:mtfnet-train}
\small
\noindent\textbf{Require:} Dataset $\mathcal{D}=\{(X_i,y_i)\}_{i=1}^N$, sequence length $T$, batch size $B$, learning rate $\eta$, weight decay $\lambda$, grad clip $c$, epochs $E$\\
\textbf{Ensure:} Trained parameters $\theta^\star$
\begin{algorithmic}[1]
\State \textbf{Initialize} parameters for kinematics, local, and scene encoders; fusion GLU; classifier; optimizer (RMSProp with lr $\eta$, weight decay $\lambda$); AMP; scheduler
\For{$\text{epoch} \gets 1 \ \text{to}\ E$}
  \ForAll{batches $(X^{(b)}, y^{(b)}) \subset \mathcal{D}$}
    \State $(P_{\text{bb}}, P_{\text{bp}}, P_{\text{lc}}, E_{\text{sc}}) \gets X^{(b)}$ \Comment{Unpack multi-modal inputs}
    \State $Z_{\text{bp}} \gets \textsc{GLU}(W_{\text{bp}} P_{\text{bp}})$
    \State $h_{\text{pose}} \gets \textsc{Attn}\!\big(\textsc{GRU}_{\text{pose}}(Z_{\text{bp}})\big)$
    \State $h_{\text{bb}} \gets \textsc{GLU}\!\big(W_{\text{bb}}\, P_{\text{bb}}[:, T, :]\big)$
    \State $h_{\text{kin}} \gets \textsc{GLU}\!\big(\textsc{Concat}(h_{\text{pose}},\, h_{\text{bb}})\big)$
    \State $h_{\text{lc}} \gets \textsc{Attn}\!\big(\textsc{GRU}_{\text{lc}}(\textsc{GLU}(\textsc{CNN}(P_{\text{lc}})))\big)$
    \State $h_{\text{sc}} \gets \textsc{Attn}\!\big(\textsc{GRU}_{\text{sc}}(\textsc{GLU}(\textsc{Conv3D}(E_{\text{sc}})))\big)$
    \State $h \gets \textsc{Concat}(h_{\text{kin}},\, h_{\text{lc}},\, h_{\text{sc}})$
    \State $\hat{y} \gets \sigma\!\big(\textsc{Linear}(\textsc{GLU}(h))\big)$
    \State $\mathcal{L} \gets \textsc{BCEWithLogitsLoss}(\hat{y},\, y^{(b)}) + \lambda \lVert W \rVert_2^2$
    \State \textsc{AMPForwardBackward}$(\mathcal{L})$; \textsc{ClipGradients}$(\lVert g \rVert_2 \le c)$; \textsc{RMSPropStep}()
  \EndFor
  \State \textsc{UpdateScheduler}(); 
  
  \textbf{if} \textsc{ValAUCPlateau}() \textbf{ then} 
  \textsc{EarlyStop}(); 
\EndFor
\State \Return $\theta^\star$
\end{algorithmic}
\end{algorithm}
\vspace{-12pt}
\subsection{Multi-Branch Temporal Encoding}
After feature extraction, MTF-Net employs three parallel temporal encoding branches 
to capture complementary behavioral cues: (i) the \textit{kinematics branch} modeling fine body motion, 
(ii) the \textit{local context branch} capturing appearance and short-range cues, 
and (iii) the \textit{scene context branch} reasoning over global spatio-temporal semantics. 
Each branch transforms its input sequence into a compact temporal embedding, 
which is later fused for holistic intention estimation. \textbf{(a) Kinematics Branch.} The kinematics branch jointly models pose dynamics and bounding-box displacement. Given the sequence of normalized pose vectors 
$P_{bp} \in \mathbb{R}^{T\times 2K}$ and bounding boxes 
$P_{bb} \in \mathbb{R}^{T\times 4}$, each frame-level pose feature is first 
projected through a GLU layer:
\begin{equation}
Z_{bp,t} = \text{GLU}\!\big( W_{bp} P_{bp,t} + b_{bp} \big) \in \mathbb{R}^{d_p}.
\end{equation}
where $W_{bp}\!\in\!\mathbb{R}^{d_p\times 2K}$ and $d_p$ are the projected dimension.
The sequence $\{Z_{bp,t}\}_{t=1}^{T}$ is then processed by a GRU
to capture temporal dependencies:
\begin{equation}
H_{bp} = \text{GRU}_{\text{pose}}(Z_{bp}) \in \mathbb{R}^{T \times d_p}.
\end{equation}
A temporal attention mechanism aggregates informative frames:
\begin{equation}
\begin{aligned}
    \alpha_t &= 
    \frac{\exp(w_a^{\top}\tanh(W_a H_{bp,t}))}
         {\sum_{i=1}^{T}\exp(w_a^{\top}\tanh(W_a H_{bp,i}))}, \\
    h_{bp} &= \sum_{t=1}^{T}\alpha_t H_{bp,t},
\end{aligned}
\end{equation}
Finally, the attended pose embedding $h_{bp}$ is concatenated with the 
last-frame bounding-box feature $P_{bb,T}$ projected by a GLU layer,
\begin{equation}
h_{\text{kin}}
= \text{GLU}\!\big( W_{\text{kin}} [\, h_{bp};\, P_{bb,T} \,] + b_{\text{kin}} \big)
\in \mathbb{R}^{d_{\text{kin}}}.
\end{equation}
yielding the kinematic representation 
$h_{\text{kin}} \in \mathbb{R}^{d_{\text{kin}}}$. \textbf{(b) Local Context Branch.}
The local context branch encodes the pedestrian-centered RGB crops 
$P_{lc}\!\in\!\mathbb{R}^{T\times 3\times H_c\times W_c}$ to model 
appearance and near-field motion cues. 
Each frame is processed by a pretrained convolutional backbone 
$g_{\text{cnn}}(\cdot)$ (VGG19~\cite{Simonyan2014VeryDC}) followed by global average pooling (GAP):
\begin{equation}
z_{lc,t} = \text{GAP}\!\big( g_{\text{cnn}}(P_{lc,t}) \big) \in \mathbb{R}^{d_v}.
\end{equation}
The visual embeddings $\{z_{lc,t}\}$ are linearly projected through a GLU,
encoded with a GRU, and summarized by temporal attention:
\begin{equation}
h_{\text{loc}}
= \text{Attn}\!\left(\text{GRU}_{\text{lc}}\!\big(\text{GLU}(z_{lc})\big)\right)
\in \mathbb{R}^{d_{\text{loc}}}.
\end{equation}
\textbf{(c) Scene Context Branch.} The scene context branch reasons over broader environmental semantics 
from $E_{sc}\!\in\!\mathbb{R}^{T\times C\times H_s\times W_s}$. 
We first permute the temporal and channel dimensions to obtain a 
5D tensor $(C,\,T,\,H_s,\,W_s)$ and apply a series of 3D convolutional layers 
to jointly capture spatial and temporal correlations:
\begin{equation}
Z_{sc} = \text{Conv3D}_{1:L}(E_{sc}) \in \mathbb{R}^{T \times d_s \times H' \times W'}.
\end{equation}
After global spatial pooling, the sequence $\{z_{sc,t}\}$ 
is passed through a GRU and attention module analogous to the other branches:
\begin{equation}
h_{\text{scene}}
= \text{Attn}\!\left(\text{GRU}_{\text{sc}}(z_{sc})\right)
\in \mathbb{R}^{d_{\text{scene}}}.
\end{equation}
\textbf{Output Representations.} Each branch produces a temporally aggregated feature vector:
\begin{equation}
\begin{aligned}
    h_{\text{kin}} &\in \mathbb{R}^{d_{\text{kin}}}, h_{\text{loc}} &\in \mathbb{R}^{d_{\text{loc}}}, h_{\text{scene}} &\in \mathbb{R}^{d_{\text{scene}}},
\end{aligned}
\end{equation}
which collectively represent kinematic, local, and global contextual information. 
These embeddings are subsequently fused in the global fusion module 
(Sec.~\ref{sec:fusion}) to produce the final intention probability.

\begin{table*}[!ht]
\centering
\caption{Quantitative comparison of state-of-the-art pedestrian intention prediction models on the PIE and JAAD benchmarks. Reported metrics include Accuracy (Acc), AUC, F1-score, Precision (Prec.), and Recall (Rec.), averaged over three independent runs.}

\begin{tabular}{l|ccccc|ccccc}
\hline
\multirow{2}{*}{\textbf{Model}} & \multicolumn{5}{c|}{\textbf{PIE Dataset}} & \multicolumn{5}{c}{\textbf{JAAD Dataset}} \\ 
\cline{2-11} & \textbf{Acc} & \textbf{AUC} & \textbf{F1} & \textbf{Prec.} & \textbf{Rec.} & \textbf{Acc} & \textbf{AUC} & \textbf{F1} & \textbf{Prec.} & \textbf{Rec.} \\ 
\hline
PIE~\cite{rasouli2019pie} (Baseline) & 0.79 & 0.74 & 0.87 & 0.86 & 0.88 & 0.77 & 0.72 & 0.85 & 0.84 & 0.86 \\ 
\hline
SingleRNN \cite{kotseruba2020they} & 0.81 & 0.75 & 0.64 & 0.61 & 0.70 & 0.78 & 0.74 & 0.62 & 0.59 & 0.68 \\
SFRNN \cite{rasouli2020pedestrian} & 0.84 & 0.82 & 0.72 & 0.75 & 0.80 & 0.82 & 0.81 & 0.71 & 0.73 & 0.78 \\
PCPA \cite{kotseruba2021benchmark} & 0.87 & 0.83 & 0.77 & 0.80 & 0.83 & 0.84 & 0.82 & 0.75 & 0.77 & 0.81 \\
CAPformer \cite{s21175694} & 0.88 & 0.80 & 0.71 & 0.69 & 0.79 & 0.85 & 0.81 & 0.70 & 0.68 & 0.77 \\
PPCI \cite{yang2022predicting} & 0.89 & 0.90 & 0.81 & 0.79 & 0.82 & 0.86 & 0.88 & 0.80 & 0.78 & 0.82 \\
GraphPlus \cite{cadena2022pedestrian} & 0.89 & 0.91 & 0.84 & 0.83 & 0.83 & 0.87 & 0.89 & 0.82 & 0.80 & 0.84 \\
MCIP \cite{ham2022mcip} & 0.89 & 0.87 & 0.81 & 0.81 & 0.83 & 0.86 & 0.85 & 0.79 & 0.79 & 0.82 \\
CIPF \cite{ham2023cipf} & 0.91 & 0.89 & 0.83 & 0.85 & 0.83 & 0.89 & 0.87 & 0.82 & 0.83 & 0.84 \\
PIT \cite{zhou2023pit}  & 0.91 & 0.92 & 0.87 & 0.86 & 0.85 & 0.90 & 0.91 & 0.86 & 0.85 & 0.86 \\
VMIGI \cite{sharma2023visual} & 0.92 & 0.93 & 0.87 & 0.87 & 0.88 & 0.89 & 0.92 & 0.87 & 0.86 & 0.87 \\
TrEP \cite{zhang2023trep} & 0.92 & 0.94 & 0.88 & 0.89 & 0.84 & 0.91 & 0.93 & 0.87 & 0.88 & 0.85 \\
PIP-Net  \cite{azarmi2025pip} & 0.92 & 0.94 & 0.88 & 0.89 & 0.88 & 0.91 & 0.93 & 0.88 & 0.89 & 0.87 \\
\cline{1-11}
\textbf{MTF-Net (Ours)} & \textbf{0.93} & \textbf{0.95} & \textbf{0.90} & \textbf{0.91} & \textbf{0.89} & \textbf{0.93} & \textbf{0.94} & \textbf{0.89} & \textbf{0.90} & \textbf{0.88} \\
\hline
\end{tabular}
\label{tab:combined_results}
\end{table*}

\subsection{Multi-Modal Fusion and Classification Head}
\label{sec:fusion}

The embeddings obtained from the three temporal branches are jointly fused to derive a unified representation of the pedestrian’s behavioral context. 
Let $h_{\text{kin}} \in \mathbb{R}^{d_{\text{kin}}}$, 
$h_{\text{loc}} \in \mathbb{R}^{d_{\text{loc}}}$, and 
$h_{\text{scene}} \in \mathbb{R}^{d_{\text{scene}}}$ 
denote the final feature vectors produced by the kinematics, local context, and scene context branches, respectively. 
These embeddings are concatenated and linearly projected into a shared latent space:
\begin{equation}
\begin{aligned}
    h_{\text{fuse}} &= [\, h_{\text{kin}};\, h_{\text{loc}};\, h_{\text{scene}} \,], \\
    h_{\text{fuse}} &\in \mathbb{R}^{\, d_{\text{kin}} + d_{\text{loc}} + d_{\text{scene}}}.
\end{aligned}
\end{equation}
To enable nonlinear interactions across modalities, a GLU fusion layer is applied:
\begin{equation}
z = \text{GLU}\!\big( W_f h_{\text{fuse}} + b_f \big) \in \mathbb{R}^{d_z}.
\end{equation}
where $W_f \in \mathbb{R}^{d_z \times (d_{\text{kin}} + d_{\text{loc}} + d_{\text{scene}})}$ 
and $d_z$ denotes the dimension of the fused embedding. 
The GLU mechanism dynamically regulates cross-modal information flow, emphasizing salient cues while suppressing redundant or weakly correlated features. The fused representation is then passed through a shallow classification head to estimate the probability of pedestrian crossing intention:
\begin{equation}
\hat{y}
= \sigma\!\big( W_c\, \text{GLU}(W_h z + b_h) + b_c \big)
\in [0,1].
\end{equation}
where $\sigma(\cdot)$ denotes the sigmoid activation function. Dropout regularization (rate 0.3–0.5) and gradient clipping ($\|g\|_2 \leq 1.0$) are employed during training to enhance stability and generalization. Following this formulation, MTF-Net is optimized end-to-end using mixed-precision RMSProp training with L2 weight decay and adaptive learning-rate scheduling. 
The overall optimization workflow, integrating multi-branch temporal encoding and GLU-based fusion, is summarized in Algorithm~\ref{alg:mtfnet-train}. 
This fusion strategy allows MTF-Net to jointly reason over motion, appearance, and environmental context, yielding a temporally coherent and semantically rich prediction of pedestrian crossing intention.

\begin{figure} [!ht]
    \centering
    \includegraphics[width=1\linewidth]{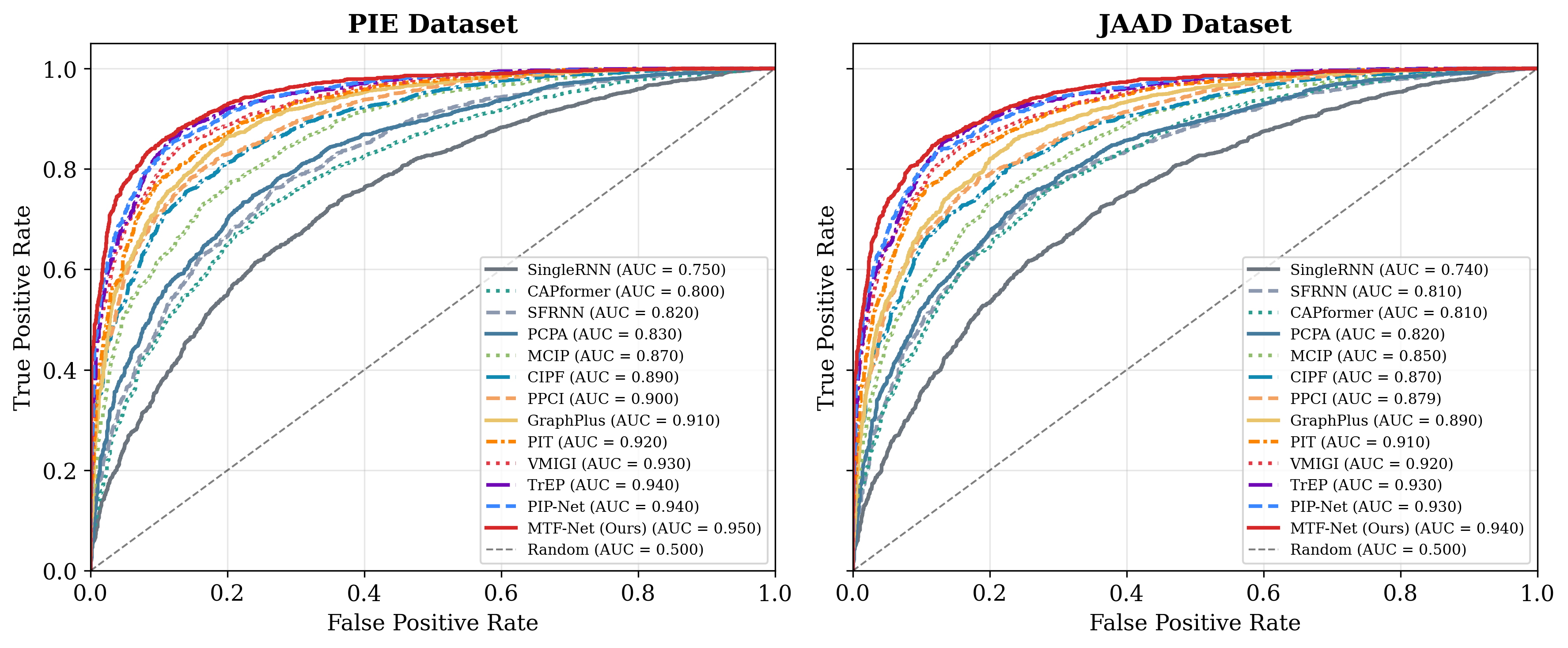}
    \caption{Comparison of Receiver Operating Characteristic (ROC) curves across pedestrian intention prediction models on the PIE and JAAD datasets. Each curve illustrates the trade-off between true positive rate and false positive rate for a given model, with the shaded area summarizing overall discriminative ability (AUC).}
    \label{fig:auc_roc_comparison}
\end{figure}
\vspace{-16pt}
\subsection{Training Objective and Optimization}
The proposed MTF-Net is trained to minimize the discrepancy between the predicted crossing probability $\hat{y}$ and the ground-truth label $y \in \{0,1\}$. 
A binary cross-entropy loss with logits is employed:
\begin{equation}
     \mathcal{L}_{\text{BCE}} 
     = -\,\big[\, y \log(\hat{y}) + (1 - y)\log(1 - \hat{y}) \,\big].
 \end{equation}
 For a dataset $\mathcal{D} = \{(X_i, y_i)\}_{i=1}^{N}$, 
 the overall objective becomes
 \begin{equation}
     \mathcal{L}_{\text{total}} 
     = \frac{1}{N} \sum_{i=1}^{N} \mathcal{L}_{\text{BCE}}(X_i, y_i)
     + \lambda \|W\|_2^2,
 \end{equation}
 where the second term denotes $L_2$ weight regularization with coefficient $\lambda$.
\section{Experimental Setup}
\subsection{Dataset}
\textbf{Pedestrian Intention Estimation (PIE) dataset~\cite{rasouli2019pie}} contains over 6.8 hours of ego-vehicle dashcam videos recorded in urban traffic environments at 30\,fps, comprising approximately 1.6$\times$10$^{5}$ annotated pedestrian bounding boxes and 1.8$\times$10$^{4}$ tracklets with corresponding crossing intention labels. Each pedestrian is annotated with spatial bounding boxes, behavior states (walking, standing, crossing), and temporal intention labels defining whether the pedestrian will cross within the next 1–4\,s. \textbf{Joint Attention for Autonomous Driving (JAAD) dataset~\cite{rasouli2017ICCVW}} includes 346 video clips (about 82{,}000 annotated frames) collected at 30\,fps, capturing urban intersections and yielding scenarios with diverse pedestrian–vehicle interactions. Both datasets are unified to a frame rate of 15\,fps and a consistent annotation structure containing bounding boxes, pose keypoints, and crossing-intention labels. Each pedestrian trajectory is segmented into observation windows of $T$ frames preceding the decision point, where $T\!\in\!\{10, 15, 20\}$ frames depending on the prediction horizon $\tau$. For PIE, we adopt the official data split of 70\% for training, 15\% for validation, and 15\% for testing, while for JAAD, sequences are split approximately 60\%/20\%/20\% for train, validation, and test, respectively. This unified preprocessing ensures consistent temporal resolution and annotation semantics across datasets, enabling fair evaluation of MTF-Net’s generalization capability.

\subsection{Evaluation Metrics}
To comprehensively assess pedestrian intention prediction performance, we employ multiple quantitative metrics that capture both classification accuracy and probabilistic reliability. For each evaluation horizon $\tau \!\in\! \{1,2,3,4\}$\,seconds, predictions are compared with ground-truth labels using Accuracy, Precision, Recall, and the F1-score, computed as the harmonic mean of precision and recall. 
To evaluate probabilistic confidence, we additionally report the Area Under the Receiver Operating Characteristic Curve (AUC), which measures the model’s discrimination ability across varying thresholds. 
Given the binary intention label $y\!\in\!\{0,1\}$ and predicted probability $\hat{y}\!\in\![0,1]$, a prediction is considered crossing if $\hat{y}\!>\!0.5$. 
Performance is averaged over all pedestrian instances and three independent runs to ensure statistical robustness. AUC is used as the primary selection criterion for model convergence and early stopping, as it provides a threshold-independent indicator of overall predictive quality.

\subsection{Implementation Details}
All experiments are implemented in PyTorch and conducted on four NVIDIA RTX~4090 GPUs (24\,GB VRAM). The model is trained using the RMSProp optimizer with an initial learning rate of $5\times10^{-5}$ and a batch size of~32. The binary cross-entropy loss with logits is employed as the training objective, incorporating $L_2$ weight regularization with coefficient $10^{-4}$. Training is performed for up to 100 epochs with early stopping based on validation AUC. A learning-rate scheduler (\textit{ReduceLROnPlateau}) reduces the rate by a factor of~0.5 upon performance stagnation. All layers utilize GLU activations, dropout rates between~0.3 and~0.5, and gradient clipping with $\|g\|_2\!\leq\!1.0$ for stable optimization. Mixed-precision training (AMP) is enabled to accelerate convergence and reduce memory consumption. During inference, the trained MTF-Net model achieves an average latency of approximately 59\,ms per sample ($\sim$17\,FPS) on a single RTX~4090 GPU, confirming its real-time capability for deployment in onboard ADAS systems.

\begin{table*}[!ht]
\centering
\caption{Feature ablation results on the PIE dataset. Progressively incorporating local context (LC) and scene context (SC) features improves all metrics, demonstrating the benefit of multi-modal fusion in MTF-Net. Bold values indicate the best results.}
\label{tab:ablation_feature}
\begin{tabular}{@{}lccccc@{}}
\toprule
\textbf{Configuration} & \textbf{AUC} & \textbf{F1} & \textbf{Accuracy} & \textbf{Precision} & \textbf{Recall} \\
\midrule
BB + Pose                  & 0.900 & 0.830 & 0.890 & 0.810 & 0.840 \\
BB + Pose + LC             & 0.930 & 0.870 & 0.910 & 0.880 & 0.860 \\
BB + Pose + LC + SC (Full) & \textbf{0.950} & \textbf{0.900} & \textbf{0.930} & \textbf{0.910} & \textbf{0.890} \\
\bottomrule
\end{tabular}
\end{table*}

\begin{figure*}[!ht]
    \centering
    \includegraphics[width=\linewidth]{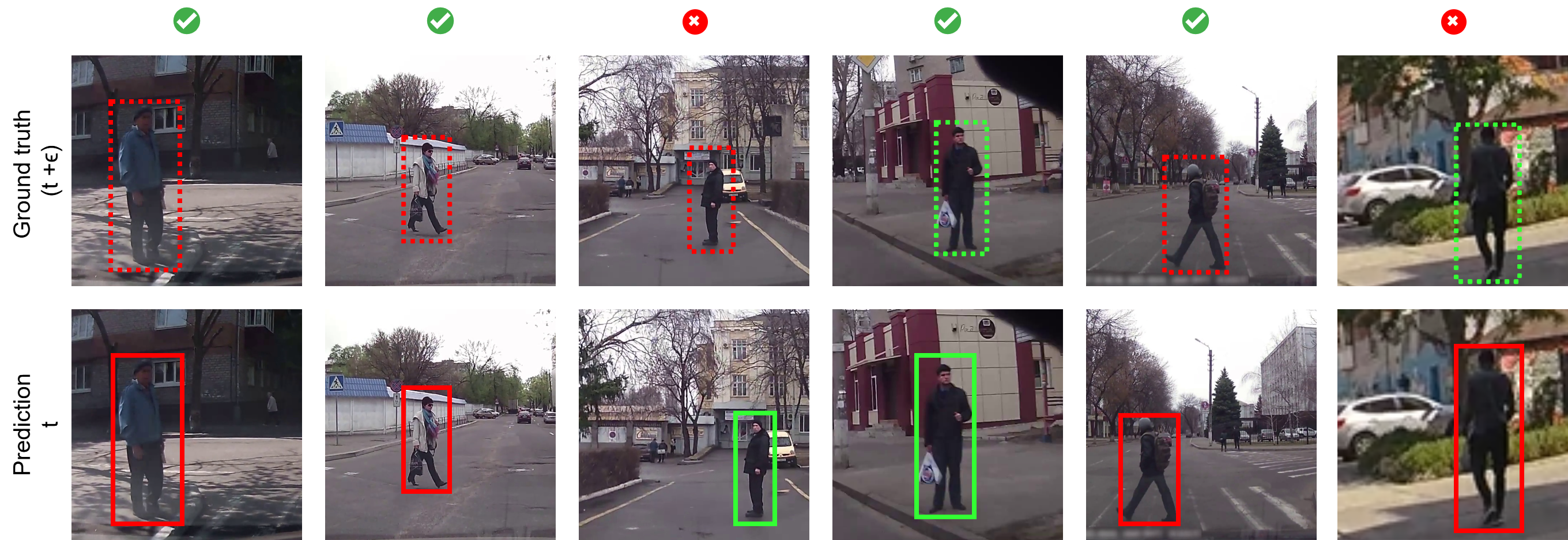}
    \caption{Qualitative visualization of pedestrian intention prediction on PIE and JAAD sequences. Bounding boxes are color-coded from green (low) to red (high) according to the predicted crossing probability, with temporal probability curves displayed below each sequence. MTF-Net accurately anticipates crossing intent several frames before it occurs, attending to key visual cues such as body motion and orientation. The frame at $t$ represents the decisive prediction moment, while $(t + \epsilon)$ denotes the corresponding ground truth.}
    \label{fig:QualitativeJAAD}
\end{figure*}
\vspace{-12pt}
\section{Results and Analysis}
\subsection{Quantitative Results}
As shown in Table~\ref{tab:combined_results}, MTF-Net achieves the best overall performance across both the PIE and JAAD benchmarks, consistently outperforming prior state-of-the-art methods on all reported metrics. On the PIE dataset, it reaches an accuracy of 0.93 and an AUC of 0.95, improving over the previous best accuracy of 0.92 and AUC of 0.94 by around 1 percentage point. Similar gains are observed in F1-score, where MTF-Net attains 0.90 compared to the strongest competing scores of 0.88 (approximately a 2\% relative improvement), alongside higher precision (0.91) and recall (0.89), each exceeding earlier methods by roughly 1-2\%. On the JAAD dataset, the proposed model again leads with 0.93 accuracy and 0.94 AUC, surpassing the closest competitors at 0.91 accuracy and 0.93 AUC by about 2\% and 1\%, respectively. The F1-score of 0.89 also slightly improves on the previous best of 0.88, while precision and recall reach 0.90 and 0.88, maintaining margins of around 1--2\% over the strongest baselines. Overall, Table~\ref{tab:combined_results} shows that MTF-Net consistently achieves small but uniform improvements across all metrics on both datasets.

\subsection{Qualitative Visualization}
Fig.~\ref{fig:QualitativeJAAD} presents qualitative examples of pedestrian intention prediction on representative sequences from the PIE and JAAD datasets. In each sequence, pedestrian bounding boxes are color-coded from green (low) to red (high) based on the predicted crossing probability, while the lower panel shows the corresponding temporal probability curve $\hat{y}_t$ aligned with the ground-truth intention label. The results show that MTF-Net anticipates crossing intention several frames before the pedestrian actually initiates the crossing action. The predicted probability increases smoothly as subtle preparatory indicators such as changes in posture, head orientation, or step initiation appear in the observation window. This early response demonstrates the model’s ability to capture fine-grained temporal dependencies and causal motion indicators rather than relying solely on instantaneous visual features.

\begin{table}[!ht]
\centering
\caption{Effect of sequence length and activation function on the PIE dataset. Results are averaged over three runs.}
\begin{tabular}{@{}lccccc@{}}
\hline
\textbf{Configuration} & \textbf{AUC} & \textbf{F1} & \textbf{Acc} & \textbf{Prec.} & \textbf{Rec.} \\
\hline
$T=10$ (short)   & 0.92 & 0.86 & 0.90 & 0.87 & 0.85 \\
$T=15$ (default) & \textbf{0.95} & \textbf{0.90} & \textbf{0.93} & \textbf{0.91} & \textbf{0.89} \\
$T=20$ (long)    & 0.93 & 0.88 & 0.91 & 0.89 & 0.87 \\
\hline
ReLU activation   & 0.94 & 0.88 & 0.92 & 0.89 & 0.87 \\
GLU activation    & \textbf{0.95} & \textbf{0.90} & \textbf{0.93} & \textbf{0.91} & \textbf{0.89} \\
\hline
\end{tabular}
\label{tab:ablation_seq}
\end{table}
\vspace{-13pt}
\subsection{Ablation Studies}
We conduct ablation experiments on the PIE dataset to evaluate the contribution of key components in MTF-Net, including (i) input modalities, (ii) sequence length~$T$, and (iii) activation type (GLU vs.\ ReLU). 
All results are averaged over three runs for reliability.

\textbf{(a) Feature Modality Contribution.}
Table~\ref{tab:ablation_feature} shows results obtained by progressively adding feature modalities: bounding box (BB), pose keypoints (Pose), local context (LC), and scene context (SC). 
Performance improves steadily as additional cues are integrated particularly LC and SC which capture fine-grained visual appearance and broader environmental semantics.  The full configuration (BB + Pose + LC + SC) achieves the highest overall accuracy and AUC, confirming the complementary nature of the four modalities.

\textbf{(b) Sequence Length and GLU Effect.}
Table~\ref{tab:ablation_seq} summarizes the influence of temporal window length~$T$ and activation choice. 
Short sequences ($T\!=\!10$) lack sufficient motion context, while overly long ones ($T\!=\!20$) add noise. 
The default $T\!=\!15$ offers the best trade-off between accuracy and temporal coverage. 
Replacing GLU with ReLU slightly reduces performance, demonstrating that GLU’s gating improves selective feature interaction across modalities.

Overall, these experiments demonstrate that multi-modal integration and GLU-based gating significantly enhance temporal reasoning and predictive stability, while a moderate observation window length ($T\!=\!15$) provides the most reliable performance.

\section{Conclusion}
\label{sec:conclusion}
In this work, we presented \textbf{MTF-Net}, an effective \textit{Multi-Modal Temporal Feature Fusion Network} for pedestrian intention prediction. The proposed architecture unifies four complementary modalities-bounding-box dynamics, pose keypoints, local context, and scene-level semantics within a recurrent, attention-guided fusion framework. By employing GLUs across temporal encoding and fusion stages, MTF-Net adaptively regulates cross-modal information flow, yielding improved interpretability and computational efficiency. Extensive experiments on the PIE and JAAD benchmarks demonstrate that MTF-Net consistently outperforms recent transformer- and graph-based baselines in both accuracy and AUC, while maintaining real-time inference and stable convergence. Beyond its empirical performance, MTF-Net establishes a principled framework for interpretable and scalable multi-modal fusion an essential capability for safety-critical applications such as autonomous driving and ADAS. In the future work, we plan to extend this framework by incorporating ego-vehicle dynamics and multi-agent social interactions to model richer behavioral dependencies. We also aim to explore vision language grounding and uncertainty-aware inference to enhance the robustness and situational awareness of intention prediction in complex urban environments.
%
%
\bibliographystyle{splncs04}
\bibliography{references}

\end{document}